# Decision-Support and Modeling with Large Language Models for Geothermal Well Arrays

Edwin Ouko[1,*], Emmanuel Lujan[1,*], Alan Edelman[1], and Robert Metcalfe[1]

[1]Julia Lab, Computer Science & Artificial Intelligence Laboratory, Massachusetts Institute of Technology

32 Vassar St, Cambridge, 02139 MA

[*]These Authors Contributed Equally to This Work

eljn@mit.edu

**Keywords:** large language models, decision-support, well arrays, coaxial wells, geothermal models, digital twins, digital multiplets, high-level high-performance programming

## ABSTRACT

Geothermal well arrays, which organize multiple geothermal wells into carefully planned geometric configurations, provide opportunities to enhance energy production capacity and increase fault tolerance. The development and adoption of these emerging geothermal technologies could be accelerated through the recent advances in large language models (LLMs) and high-level high-performance languages. A challenge in LLM-based applications is the reliability of the generated outputs, as they can be prone to subjective biases and "hallucinations". This study assesses the potential of cutting-edge LLMs—such as ChatGPT, Gemini, Claude, Grok, and domain-specific models like AskGDR— as expert assistants that can synthesize insightful interpretations of complex geothermal data, as well as improve feature capabilities of geothermal models and numerical software. We developed a novel approach, leveraging Google's recently introduced AI assistant, NotebookLM, to accelerate the generation of unpublished quantitative geothermal benchmarks. The rapid generation of these evaluation instruments is essential for assessing the swiftly evolving capabilities of emerging language model technologies. In particular, we use these benchmarks and LLM-based interviews to analyze opportunities and limitations of two promising technologies: geothermal well arrays and closed-loop coaxial wells. Furthermore, we present a case study illustrating how LLMs can facilitate auto-parallelization of geothermal numerical models. Our analysis emphasizes their application in digital twins and underscores the importance of high-level, high-performance code generation. This line of research could play a transformative role in the geothermal sector by enabling the next-generation of decision-support applications, integrating data analysis, informed recommendations, and more dynamic numerical modeling workflows.

## 1. INTRODUCTION

### 1.1 LLMs for Geothermal Well Arrays and Closed-loop Coaxial Wells

World leaders at COP28[1] agreed to "tripling renewable energy capacity globally" to 11,000 gigawatts by 2030[2], a crucial step in addressing climate change. The use of fossil fuels for power generation is one of the leading contributors to global warming (U.S. Department of Energy, Office of Energy Efficiency & Renewable Energy). Despite the abundance of renewable energy sources, reliability remains a significant concern, as many are intermittent or have limited production capacities (Notton et al., 2018). This challenge has heightened interest in geothermal energy as a more dependable alternative. Emerging geothermal technologies, including well arrays and closed-loop coaxial wells, present a promising solution to the critical challenge of generating cleaner and more economically viable energy. Geothermal well arrays, as depicted in Figure 1, cluster multiple geothermal wells in strategic geometric patterns, and offer opportunities to boost energy production capacity and improve fault tolerance (Ding and Wang et al., 2018). These arrays can be composed of coaxial closed-loop geothermal systems, a technology that could substantially reduce the cost of drilling (Wood et al., 2012). In these systems a working fluid, such as water, is pumped into a closed subsurface conducting tube in contact with hot bed rocks, gets heated as it flows through the tube, and the heated fluid which exits through an outlet is harnessed for power generation or other uses (Liu and Dahi Taleghani et al., 2023). The coaxial design is a pipe-in-pipe system in which a cold working fluid is pumped into the well through the space between the wall of the outer pipe and the wall of the inner pipe, while the heated working fluid is simultaneously forced to the surface through the annulus by the same pump pressure (Budiono et al., 2022).

The development and adoption of novel geothermal energy technologies requires highly skilled personnel and labor-intensive efforts dedicated to analyzing complex geothermal data and generating informed recommendations. Furthermore, this process often requires geothermal modeling, a vital tool for analyzing and optimizing energy production strategies. The development of traditional software models for physical processes is both resource-intensive and reliant on the combined expertise of software engineers and domain specialists (Bryant et al., 2010).

Advances in large language models (LLMs), recognized for their powerful natural language processing capabilities, are significantly impacting various sectors, such as education, healthcare, and engineering (Chang et al., 2024). Even in the field of mathematics, a traditionally conservative field, the potential of LLMs to contribute to new theorem development is being recognized. Renowned mathematician Terence Tao examines this potential in detail (Romera-Paredes et al., 2024)[3]. In the geosciences, researchers have recently started investigating the utility of these models for data analysis, simulation, and decision support, uncovering their potential to transform traditional practices (Hadid et al., 2024). Presently, it remains unclear how LLMs will impact the field of geothermal energy, and there is a notable absence of adequate strategies or methodologies to incorporate these models into decision-support applications. Nonetheless, LLMs could play a transformative role in the geothermal sector.

[1] https://unfccc.int/cop28

[2] https://www.rff.org/publications/reports/global-energy-outlook-2024

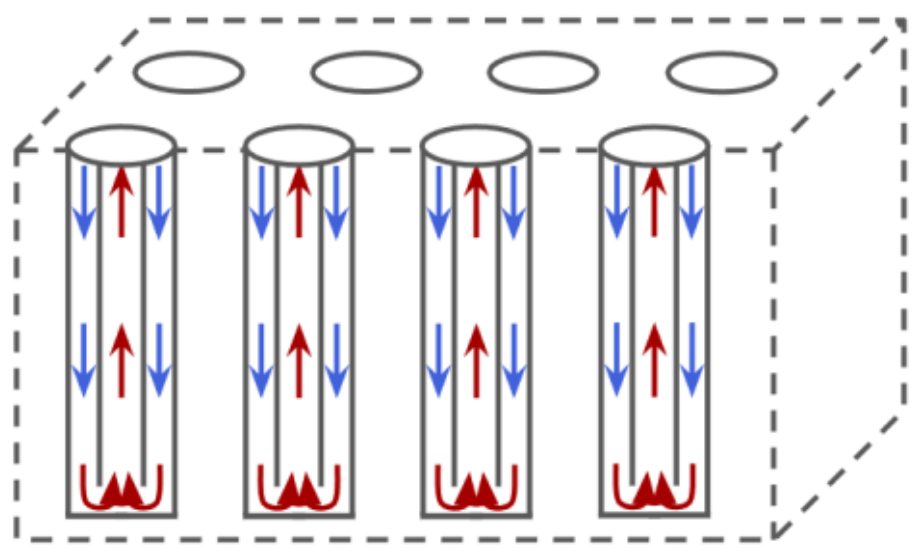

**Figure 1: Simplified scheme of a geothermal well array featuring multiple closed-loop coaxial wells.**

This section introduces LLMs in the advancement of geothermal energy through two key applications: the first focuses on assisting decision-makers by leveraging geothermal data to offer informed answers. The second examines LLMs to enhance geothermal models, with a particular emphasis on extending the capabilities of geothermal digital twins.

### 1.2 LLMs for Geothermal Data Analysis and Recommendation

The first specialized LLM designed exclusively for geoscience was K2 (Deng et al., 2024). Built on a foundation of geoscience-specific data and fine-tuned with curated instruction datasets, K2 addresses the limitations of general-purpose models in handling domain-specific queries (Deng et al., 2024). Another notable advancement in this area is the U.S. Department of Energy's Geothermal Data Repository (GDR), which now features the "AskGDR" AI research assistant (Weers et al., 2024). This tool leverages machine-readable metadata and retrieval-augmented generation capability to provide researchers with nuanced answers to complex queries. Unlike traditional keyword searches, AskGDR allows users to explore datasets in depth, accessing details about methodologies, assumptions and contextual relevance. By adhering to FAIR (Findable, Accessible, Interoperable, Reusable) and FARR (AI-readiness and Reproducibility) principles, the platform aims to enhance the accessibility and reliability of geothermal data (Weers et al., 2024).

One of the primary challenges in LLM synthesized answers is measuring reliability as LLMs are prone to hallucinations (Rawte et al., 2023). Designing benchmarks that can evaluate the quality of LLM answers beyond a simple binary framework continues to impede thorough assessment due to inherent subjectivity in human evaluations (Smith et al., 2022). One approach that has been proposed to tackle this challenge involves using multiple-choice question benchmarks to provide a more objective measure of accuracy. The GeoBench evaluation framework, developed alongside K2, provides a first assessment of the model's performance across geoscience tasks through over 1,500 objective and 939 subjective questions. This initiative sets a new standard for applying LLMs in specialized scientific domains (Deng et al., 2024). A different study focusing on geosystems' risk and uncertainty highlighted how structured prompting improves their effectiveness. By tailoring questions with additional context, researchers significantly enhanced the accuracy and reproducibility of model-generated responses, underscoring the potential of LLMs to support decision-making in geothermal projects (Mahjour et al., 2024).

Despite their success, the recent exponential growth of LLMs, such as last versions of ChatGPT, Gemini, Claude, and Grok, left decision-makers with outdated benchmarks for geothermal energy applications. In particular, to the best of our knowledge, no standardized benchmark currently exists for the most recent LLMs, and AI research assistants specialized in geothermal energy, like AskGDR. Accurate LLM-based insights can play a pivotal role in guiding decision-making for emerging or underexplored geothermal technologies.

### 1.3 LLMs for Geothermal Modelling: From Digital Twins to Digital Multiplets

Digital twins are defined as high-fidelity virtual counterparts of physical systems, sharing data in real-time to enable seamless monitoring, analysis, and optimization. These models integrate enabling technologies like Internet of Things (IoT), artificial intelligence (AI), and 5G to replicate dynamics and firmware for operational insights and predictive analyses, reshaping industries and driving innovations in areas like smart manufacturing and energy systems (Mihai et al., 2022). Digital twins impact the geothermal energy sector by enhancing real-time monitoring and predictive maintenance. Figure 2a illustrates the interaction between a physical coaxial well (Wang et al., 2020; Chmielowska et al., 2020) and its corresponding digital twin. These virtual models enable precise modeling and fault detection in geothermal drilling systems, minimizing disruptions and lowering operational costs by detecting anomalies like drill bit wear or pipe blockages early (Osinde et al., 2019). They also support high-fidelity coring for accurate subsurface analysis, crucial for understanding geological structures and mitigating risks in deep-earth explorations (Yu et al., 2020). Moreover, integrating digital twins with AI and IoT facilitates intelligent energy management, optimizing geothermal heat pumps and enhancing renewable energy systems' efficiency (Agostinelli et al., 2020).

[3] https://www.youtube.com/watch?v=_sTDSO74D8Q

The integration of LLMs into digital twins is rapidly advancing, unlocking new possibilities across various domains. In data centers, systems like ChatTwin utilize LLMs to automate 3D scene descriptions, facilitating efficient modeling of complex systems (Li et al., 2023). Explainable LLMs further enhance decision-making in dynamic digital twins by incorporating human interaction and domain-specific insights, as demonstrated in agriculture and other real-time applications (Zhang et al., 2024). In this vein, Phinyx AI[4] proposes using LLM prompts to generate multi-physics simulations—including fluid dynamics, structural analysis, and heat transfer—thereby reducing manual effort and streamlining design processes. While the combination of LLMs and numerical modeling, particularly digital twins, await comprehensive validation in real-world settings, it promises a radical change in the modeling process.

LLMs could facilitate the creation and interconnection of multiple digital twins, that we call "digital multiplets". Each digital model can work in parallel or complementarily, providing a more comprehensive understanding and a wider range of functionalities than a single digital twin, such as enabling mutual validation and cross-verification. This approach could enable the incorporation of new equations to model additional physical and chemical dynamics, adjusting the geometry and parameters for scenario-specific applications or optimize computations through model reduction or parallelization. Figure 2b illustrates the development of two variants from the original digital twin model: one modifies the geometry from a coaxial to a U-shaped well, and the other is a parallelized version to improve computational efficiency and optimize cost functions.

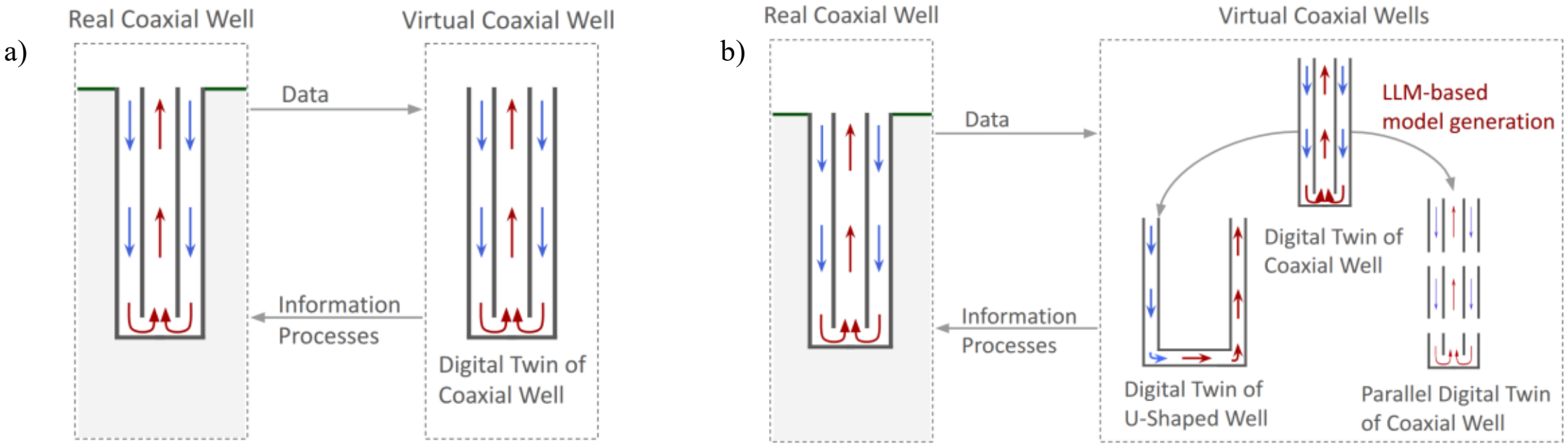

**Figure 2: Simplified scheme of a coaxial well and its digital twin (a), and scheme of a coaxial well and its digital multiplets (b).**

Another crucial aspect in the development and enhancement of geothermal numerical models, including digital twins and multiplets, is the constraints imposed by programming languages. High-level languages boost development productivity but often sacrifice performance, while high-performance languages deliver speed at the cost of ease of use. Recent advancements in scientific computing aim to bridge this gap by integrating high-level abstractions with efficient execution. A leading technology in this space is the Julia programming language—a modern, high-level, high-performance language that also supports differentiable programming for rapid prototyping and experimentation (Bezanson et al., 2012). The use of these kinds of languages may boost the productivity of LLMs in generating highly efficient geothermal codes.

To advance the development and adoption of innovative geothermal clean energy technologies, this study evaluates the capabilities and limitations of state-of-the-art LLMs for decision support. Specifically, key contributions of this article include:

- A novel approach that utilizes Google's NotebookLM, an advanced AI research tool, to accelerate the development of new LLM benchmarks tailored to geothermal energy. We use this approach to generate quantitative benchmarks for various LLMs—including ChatGPT, Gemini, Claude, and Grok—as well as the geothermal-specific model AskGDR. Building on pioneering multiple-choice question studies by (Deng et al., 2024) and (Mahjour et al., 2024), our benchmarks incorporate specialized questions and answers focused on geothermal energy, well arrays, and closed-loop coaxial wells. Additionally, LLM-based interviews were conducted to extract insights from different models on these key geothermal topics.
- A case study, where we investigate the potential of LLMs to enhance geothermal numerical models, with particular emphasis on enhancing digital twins. Specifically, we assess the auto-parallelization capabilities of different LLMs when applied to an in-house coaxial well model, and the implications of high-level, high-performance code generation.

This article is structured as follows: the Methodology section outlines the process for creating the LLM geothermal benchmarks and conducting the LLM-based auto-parallelization experiments. The Results and Discussion section evaluates the accuracy of each LLM in addressing the benchmark questions and provides a detailed analysis of the findings, including insights from a series of LLM-based interviews. Finally, the Conclusions section summarizes the key insights.

[4] https://www.phinyx.ai

## 2. METHODOLOGY

### 2.1 LLMs Reliability on Geothermal Data Analysis and Recommendation

Previous studies have tackled the challenge of measuring LLMs accuracy using both qualitative and quantitative methods. Qualitative approaches, such as evaluating the volume of generated text or the level of detail in responses via human reviewers, are often limited by the inherent subjectivity of the reviewers. In contrast, the present work adopted a more quantitative approach centered on multiple-choice questions. As noted in earlier sections, this method was first introduced by (Deng et al., 2024) and (Mahjour et al., 2024) in the geothermal field. In order to deliver a more up-to-date assessment, we evaluated state-of-the-art LLMs on a new set of questions on geothermal energy. Specifically, we considered ChatGPT o1, Gemini Advanced, Claude 3.5 Sonnet, Grok 2, alongside the recently introduced geothermal-focused generation tool, AskGDR.

To this end we introduce a novel approach to accelerate the development of LLM benchmarks, specifically tailored to geothermal research, leveraging the AI-powered research assistant, Google NotebookLM. This tool learns from user-uploaded multimodal information—such as documents, images, and websites—to help users organize and understand their data (Dihan et al., 2025). Users can interact with the AI to ask questions, receive explanations, and generate summaries, enhancing their comprehension and engagement with the material. A key feature of this AI tool is that the generated text contains citations to the sources uploaded by the user. In our study, we curated a collection of 20 academic articles focusing on general geothermal energy, geothermal closed-loop systems (including coaxial wells), and well arrays. These articles were uploaded to NotebookLM, enabling the AI assistant to synthesize the information and develop a comprehensive set of multiple-choice questions and corresponding answers. This approach significantly reduces the time required to create benchmarks. This process produces a notebook containing the generated questions, corresponding answers, and relevant source citations. We developed a microquiz for each of the specified categories. In total, we used 90 multiple choice questions for our benchmark and the number of questions in each microquiz varied between 15 and 30 inclusive.

### 2.2 LLMs-Enhanced Geothermal Models

Geothermal modeling—particularly through digital twins or multiplets—plays a crucial role in the study and optimization of energy production strategies. The inherent mathematical and computational complexity of these models often results in high computational cost functions, leading to suboptimal solutions. To mitigate the computational demands of executing geothermal models, we assess the auto-parallelization capabilities of various LLMs using an in-house closed-loop coaxial model. Instead of prioritizing physically realistic predictions, this experiment focuses on the software aspect, specifically assessing whether the generated code is genuinely parallel, accurate, and compilable. Our geothermal model is based on well-established diffusion-convection approaches (Song et al., 2018). Specifically, we implemented a full 3D representation of the heat equation, which can be expressed in a simplified form as follows:

$$\rho\, c \frac{\partial \phi}{\partial t} = \lambda \frac{\partial^2 \phi}{\partial x^2} + \lambda \frac{\partial^2 \phi}{\partial y^2} + \lambda \frac{\partial^2 \phi}{\partial z^2} - \left( \varepsilon\, vx \frac{\partial \phi}{\partial x} + \varepsilon\, vy \frac{\partial \phi}{\partial y} + \varepsilon\, vz \frac{\partial \phi}{\partial z} \right) + S$$

In this equation, **ρ** represents the density, **c** is the specific heat capacity, and **λ** denotes the thermal conductivity. The variable **φ** corresponds to temperature, while **vx**, **vy**, and **vz** are the fluid velocities in the x-, y-, and z-directions, respectively. The term **ε** is the porosity, and **S** represents the heat source within the fluid. The variable values in the equation depend on the represented sub-domains: working fluid, rock formation, and coaxial well. Convection is not modeled in the rock formation or well sub-domains, and velocities are treated as fixed values. Numerical computations were performed using the finite difference method on a uniform mesh. The Julia code developed for this implementation can be accessed on GitHub at the following link [5].

## 3. RESULTS AND DISCUSSION

### 3.1 LLMs Accuracy of Geothermal Energy Q&A

As LLMs evolve at an exponential pace, the demand for rigorous evaluation frameworks becomes increasingly important. This subsection assesses the accuracy of synthesized responses from state-of-the-art LLMs on geothermal energy topics. We introduce a novel benchmark developed using our NotebookLM-based approach and analyze the corresponding results.

A common challenge in LLM benchmarks is test set contamination, which arises when models are trained on publicly available data that is later used for evaluation (Deng et al., 2024). Moreover, preexisting questions and answers available online are often heavily restricted by copyright. Addressing these issues requires the formulation of new questions and answers. While experts in geothermal energy could design novel benchmarks that are not directly derived from widely accessible online sources, this process is time-intensive, demands specialized expertise, and incurs significant economic costs. To alleviate this burden, we present a focused seminal case study that demonstrates how LLMs can be used to streamline the generation of unpublished benchmarks. We propose a new approach leveraging the recently introduced Google AI assistant, NotebookLM. Details about the new approach are presented in the Methodology section.

[5] https://github.com/BobMetcalfe/GEO

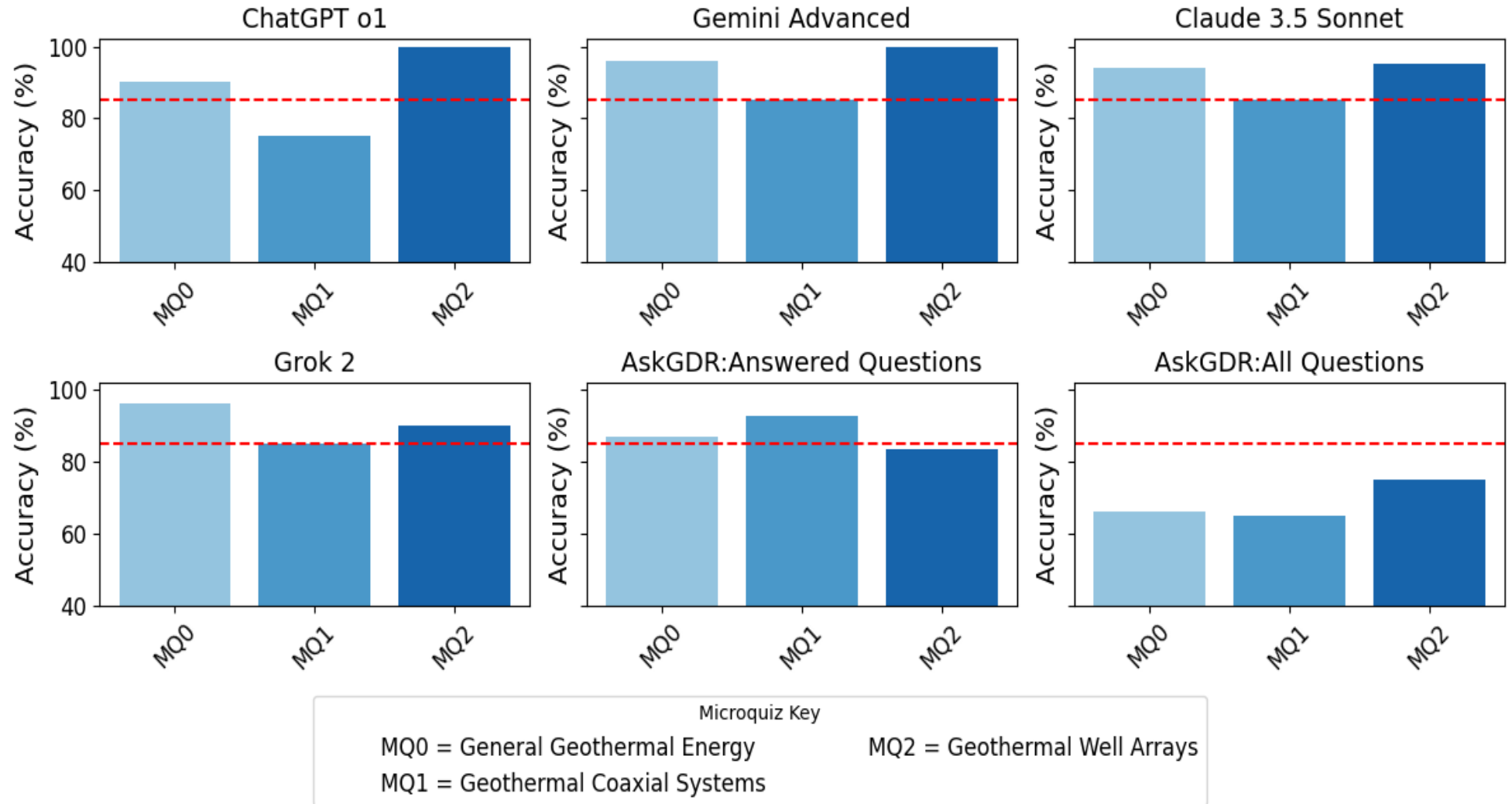

**Figure 3: Accuracy results of state-of-the-art LLMs in answering multiple-choice questions on geothermal energy based on our NotebookLM approach.**

In Figure 3, we present the results of the new generated benchmark. We display the accuracy results of the analyzed LLMs—ChatGPT o1, Gemini Advanced, Claude 3.5 Sonnet, Grok 2, and AskGDR— in answering multiple-choice questions specifically designed for geothermal energy. The questions and corresponding answers are grouped in the following categories: general geothermal energy, geothermal well arrays, geothermal closed-loop systems and coaxial wells. Given the focus of this study, this benchmark is designed to be concise while maintaining relevance. Among the analyzed models, Gemini, Claude and Grok emerged as the most reliable options, defined here as the models exhibiting the highest minimum accuracy. In particular, the accuracy threshold obtained in this benchmark is 85% which is highlighted in the figures with a dashed red line. Notably, AskGDR often responded with "I do not know" instead of guessing. To calculate AskGDR accuracy, we proceeded in two ways: in the first approach, we considered only questions where AskGDR selected one of the provided choices while in the second approach, we treated "I do not know" as a wrong answer. The first approach avoids penalizing the LLM for honesty, consistent with the factual accuracy framework widely adopted in LLM evaluation (Wei et al., 2024) while the second approach penalizes the model for not answering as many questions in the quiz as possible. Additionally, AskGDR frequently cited related sources from the geothermal data repository, even when its answers were incorrect. This feature could support the verification of responses, offering a potential avenue for ensuring answer reliability.

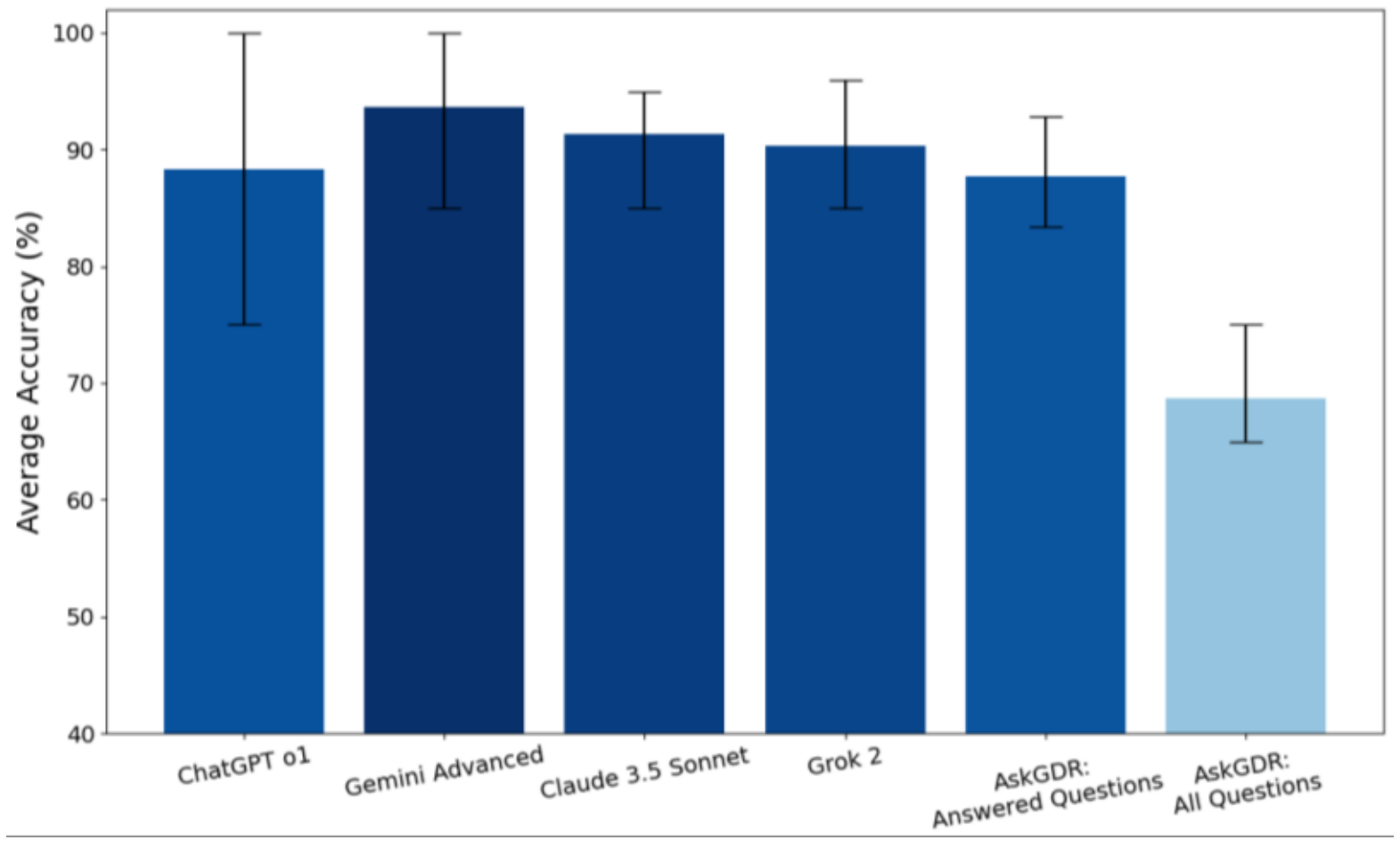

**Figure 4: Comparison of the mean, minimum, and maximum accuracies of various LLMs across all geothermal energy benchmarks.**

Figure 4 presents a comparison of the mean, minimum, and maximum accuracies across all benchmarks for the evaluated LLMs. Notably, the mean accuracy remains relatively consistent across the models, regardless of the question-and-answer category, except for ASKGDR when evaluated with all questions. However, significant differences emerge in the minimum and maximum accuracies.

Future research can build upon our approach to develop more rigorous LLM-based methodologies for analyzing the opportunities and challenges of emerging geothermal technologies. Key next steps for refining these benchmarks include expanding the sample size, validating questions and answers through expert review, and enhancing coverage across diverse geothermal energy subcategories.

Additionally, future work should focus on designing questions that assess varying levels of complexity, from foundational principles to advanced technical insights. In line with this, another valuable approach would be incorporating open-ended questions alongside multiple-choice questions and validating responses against human-annotated answers to mitigate LLM issues such as selection bias and random guessing (Myrzakhan, Bsharat, & Shen, 2024).

**3.2 LLMs-Enhanced Geothermal Models: LLM-based Auto Parallelism**

LLMs are increasingly being utilized for code development, and their capabilities can be leveraged in geothermal energy modeling, particularly for enhancing numerical digital twins or multiplets. This subsection presents our closed-loop coaxial well model and examines the capabilities of different leading LLMs for synthesizing parallel code.

In Figure 5, the temperature distribution of our closed-loop coaxial well is presented. The outer pipe injects cold fluid, while the inner pipe recovers the heated fluid at the surface. Figures 5a and 5b illustrate the initial temperature distribution and the temperature variation throughout the entire 3D domain after the working fluid has undergone heating. Figures 5c and 5d illustrate the temperature distribution at identical time points within a cross-section of the well's bottom part. As observed, the temperature outside the coaxial well has decreased, forming a green halo that indicates a cooling effect on the surrounding rock. This model serves primarily as a testbed for the auto-parallelization experiments outlined below, emphasizing this function over achieving physically realistic predictions.

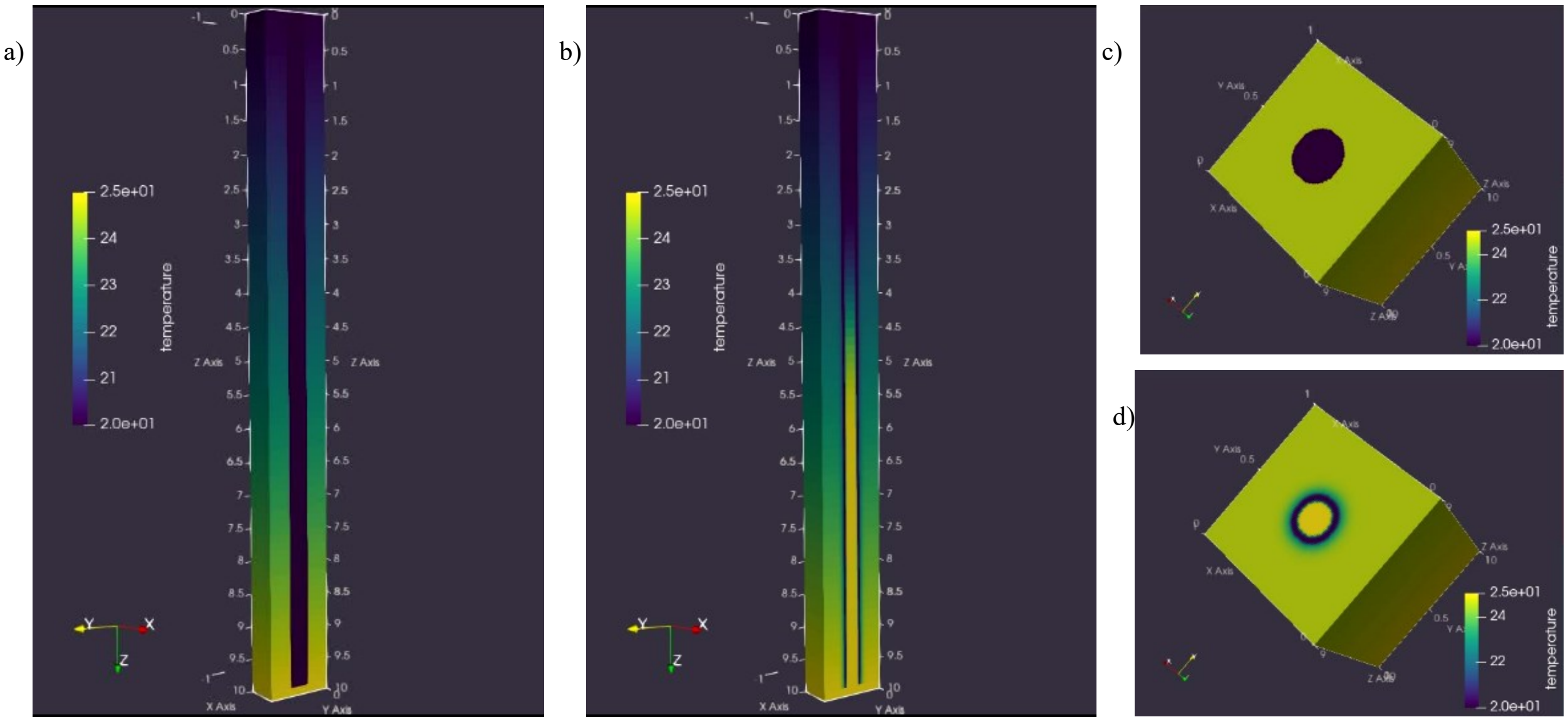

**Figure 5. Temperature distribution in the closed-loop coaxial well. Figure 5a and 5b, shows the initial temperature distribution and its evolution across the 3D domain as the working fluid undergoes heating. Figure 5c and 5d, displays the temperature distribution at the same time steps, shown in a cross-section of the well's lower region, highlighting the thermal interaction with the surrounding rock. The model serves as a testbed for the auto-parallelization experiments discussed in this subsection.**

Figure 6 presents the evaluation of the auto-parallelization capabilities of ChatGPT-o1 and Claude 3.5 Sonnet. Figure 6a illustrates the parallelization process, where the prompt "*Parallelize the Julia code below using CPU multi-threading*" was provided, followed by a serial version of our in-house coaxial well model code. Figure 6b presents the synthesized code results, showing that both LLMs accurately identified the appropriate function and line for applying the @threads macro. This modification transforms the original code into a parallel, compilable, and executable version without errors.

Gemini Advanced also parallelized the code; however, it applied the @threads macro to all loops, leading to an excessive number of spatial domain partitions. This significantly reduced the computational workload per worker, resulting in high parallelization overhead and an inefficient parallelization strategy for this case.

The LLM-based auto-parallelization approach presents significant opportunities for optimization. Traditional loop-based implementations often suffer from suboptimal cache performance. A promising direction for future research involves employing cache-oblivious divide-and-conquer stencil codes, which can achieve substantially higher efficiency but pose greater implementation challenges (Tang et al., 2011). Additionally, high-level, high-performance programming languages like Julia open new research avenues for exploring the potential of language models in generating parallel and portable code based on advanced high-performance software abstractions. Another potential experiment could focus on parallelizing geothermal simulations using task-based data-dependency parallelism. A notable example of such an approach is Dagger.jl [6] (Alomairy et al., 2024; Samaroo et al., 2024), a framework that facilitates parallel computing across diverse resources—including CPUs, GPUs, multiple threads, and distributed systems.

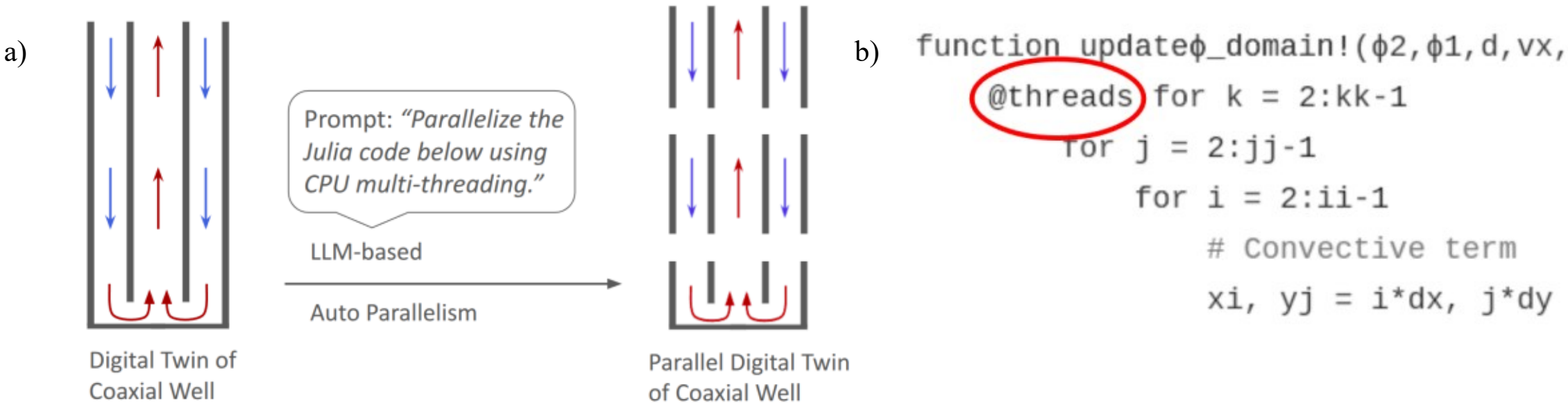

**Figure 6. Evaluation of the auto-parallelization capabilities of ChatGPT-o1 and Claude 3.5 Sonnet. Figure 6a Illustrates the parallelization process applied to a serial coaxial well model. Figure 6b displays a code fragment showing that both LLMs correctly identified where to apply the @threads macro, successfully converting the code into a parallel, compilable, and executable version without errors.**

### 3.3 LLM-Assisted Design and End-to-end Cost Modeling for Geothermal Arrays: Strengths and Limitations

The results presented in the previous subsections indicate that LLMs achieve a minimum accuracy of 85% when responding to diverse queries related to geothermal energy. The continuous refinement of these benchmarks remains essential to facilitate the effective deployment of these advanced technologies for decision support. In this subsection, we present a series of unstructured interviews conducted with different LLMs, aimed at assessing how to scale geothermal energy production using geothermal arrays. The wells utilized in this array design are closed-loop coaxial wells, one of several possible configurations. The interview examines a 10×10 array consisting of 100 geothermal wells, numbers based on previous installations. While evaluating the validity of the predictions generated by LLMs falls beyond the scope of this study, the addressed experiments can contribute to a broader understanding of the current applicability and limitations of such AI-driven assessments.

ChatGPT and Gemini were leveraged for end-to-end brainstorming on the design of coaxial closed-loop geothermal wells and arrays[7]. The LLMs suggested several cost-saving strategies, including modular assembly, just-in-time delivery, and networked redundancy. For example, ChatGPT o1 outlined the trade-offs between different well parameters, as shown in Figure 7, but without citing supporting references.

- **Smaller Diameter**:
  - **Pros**:
    - Lower drilling cost.
    - Higher surface area-to-volume ratio, which can improve heat exchange efficiency.
  - **Cons**:
    - Lower flow rate, potentially limiting thermal output.
    - Higher fluid velocity, which can increase pressure drop and pumping cost.
- **Closer Spacing (~20–40 meters)**:
  - **Pros**:
    - Allows more wells in a given area, increasing total power output per field.
  - **Cons**:
    - Higher risk of thermal interference between wells, reducing long-term efficiency.

**Figure 7: ChatGPT o1 Showing the Tradeoffs Associated with Different Diameters and Array Well Spacings**

While LLMs offer valuable support for design and cost modeling, their limitations must be addressed to maximize their utility. A significant challenge is the lack of references in responses, making it difficult to verify claims and mitigate hallucinations. Some tools, such as Perplexity[8], have taken steps to address this issue. However, LLMs like ChatGPT and Gemini often fail to cite the sources of data used in calculations, key facts, and underlying assumptions (Sanchez-Ramos et al., 2023). This absence of citations necessitates external validation of the information provided. Figure 8 presents an example of a generated response in which ChatGPT fails to disclose its underlying assumptions and the sources of the numerical data used.

Another overarching limitation of using LLMs is the reliance of the quality of the response on the prompter's ability to provide well-stated questions with a lot of context in the prompts (Ekin, 2023).

---

[6] https://github.com/JuliaParallel/Dagger.jl

[7] https://github.com/BobMetcalfe/GEO

[8] https://www.perplexity.ai

After applying learning curve reductions, the **Levelized Cost of Energy (LCOE)** for the geothermal array is re-estimated at **$44.53 per MWh**.

This is a significant improvement from the initial LCOE of **$87.90 per MWh**, making the array much more competitive with even the lowest-cost renewable energy sources like solar and wind, while maintaining the advantage of consistent baseload power.

Would you like to explore further cost reductions, sensitivity analysis, or deployment scenarios?

**Figure 8: An Example of an LLM Failing to State the Origin of the new LCOE**

The application of LLMs in geothermal design brainstorming exhibits considerable potential, particularly for individuals seeking to expand their expertise in the field or explore innovative design approaches. These models can generate reasonable optimization strategies and effectively outline key considerations for geothermal facilities and wells. However, to serve as a foundational framework for the design process or decision support, AI-generated insights must be reinforced with robust validation mechanisms

## 4. CONCLUSION

The overarching long-term goal of this interdisciplinary study is to accelerate the adoption of emerging clean geothermal technologies, including geothermal well arrays and closed-loop coaxial wells, to enhance clean energy production while minimizing economic costs. To accomplish this, we leverage the latest advancements in LLM technologies combined with state-of-the-art high-level, high-performance languages.

This study assessed the accuracy of geothermal energy responses generated by leading general-purpose LLMs—ChatGPT o1, Gemini Advanced, Claude 3.5 Sonnet, Grok 2, and AskGDR—as well as the geothermal-specific model, AskGDR. To facilitate the creation of geothermal benchmarks tailored to LLM-generated answers—a necessity that is expected to grow alongside the rapid expansion of these models—we introduced an approach leveraging Google AI's NotebookLM. We presented a seminal case demonstrating how LLMs can be used to evaluate other LLMs in the context of geothermal energy. Our multiple-choice benchmark results indicate that the analyzed versions of Gemini, Claude and Grok provided the most reliable responses, achieving a highest minimum accuracy of 85%. Overall, while all models demonstrated consistent mean accuracy, they differed in result dispersion.

Furthermore, we investigated the integration of LLMs with high-level, high-performance generated code as a fundamental component of geothermal numerical modeling workflows. Our case study focused on utilizing LLMs to parallelize an in-house coaxial well model. ChatGPT-o1 and Claude 3.5 Sonnet accurately identified the optimal placement for the CPU threading macro. The generated code modification produced a parallelized, error-free, and executable program. In contrast, while Gemini Advanced also implemented parallelization, it applied threading indiscriminately to all loops, resulting in excessive spatial partitions and an inefficient parallelization strategy. The integration of LLMs with high-level, high-performance synthesized code has the potential to transform geothermal numerical simulations into more dynamic and adaptable workflows, enabling faster prototyping and innovation. We anticipate that this research direction will drive the development of various software applications for decision support, including geothermal digital twins, multiplets, LLM agents, GPTs, and cloud-based platforms for numerical simulation prototyping.

## ACKNOWLEDGEMENT

This material is based upon work supported by the U.S. National Science Foundation under award Nos CNS-2346520, PHY-2028125, RISE-2425761, DMS-2325184, OAC-2103804, and OSI-2029670, by the Defense Advanced Research Projects Agency (DARPA) under Agreement No. HR00112490488, by the Department of Energy, National Nuclear Security Administration under Award Number DE-NA0003965 and by the United States Air Force Research Laboratory under Cooperative Agreement Number FA8750-19-2-1000. Neither the United States Government nor any agency thereof, nor any of their employees, makes any warranty, express or implied, or assumes any legal liability or responsibility for the accuracy, completeness, or usefulness of any information, apparatus, product, or process disclosed, or represents that its use would not infringe privately owned rights. Reference herein to any specific commercial product, process, or service by trade name, trademark, manufacturer, or otherwise does not necessarily constitute or imply its endorsement, recommendation, or favoring by the United States Government or any agency thereof. The views and opinions of authors expressed herein do not necessarily state or reflect those of the United States Government or any agency thereof." The views and conclusions contained in this document are those of the authors and should not be interpreted as representing the official policies, either expressed or implied, of the United States Air Force or the U.S. Government.

## DATA AVAILABILITY

See https://github.com/BobMetcalfe/GEO